%% file: gram2vec.tex
\documentclass[10pt, a4paper]{article}

\usepackage[final]{lrec2026} 
\usepackage{booktabs}
\usepackage{hyperref}
\usepackage{alltt}
\usepackage[T2A,T1]{fontenc}
\usepackage[utf8]{inputenc}
\usepackage[main=english,russian]{babel}
\usepackage{booktabs}
\usepackage{tcolorbox}
\usepackage{multirow}

\usepackage{xspace} 

\definecolor{pale-yellow}{HTML}{F5F176}

\usepackage{xcolor}
\usepackage{soul} 

\sethlcolor{pale-yellow}

\newcommand{\highlight}[1]{\hl{#1}} 
\newcommand{\gtv}{{Gram2Vec}\xspace}

\usepackage{todonotes}

\definecolor{peter}{rgb}{1.0, 0.6, 0.4}
\definecolor{owen}{rgb}{0.55, 0.71, 0.0}

\title{Gram2Vec: An Interpretable Document Vectorizer}

\name{Peter Zeng$^{*\ddagger}$ Hannah Stortz$^{\dagger}$ Eric Sclafani$^{\dagger}$ Alina Shabaeva$^{\dagger\ddagger}$\\
      \large \bfseries Maria Elizabeth Garza$^{\dagger}$ Daniel Greeson$^{\dagger}$ Owen Rambow$^{\dagger\ddagger}$}

\address{$^*$Department of Computer Science $^\dagger$Department of Linguistics \\ $^\ddagger$Institute for Advanced Computational Science \\
         Stony Brook University \\
         pezeng@cs.stonybrook.edu\\}

\abstract{
We present \gtv, a grammatical style embedding system that embeds documents into a higher dimensional space by extracting the normalized relative frequencies of grammatical features present in the text. Compared to neural approaches, \gtv offers inherent interpretability based on how the feature vectors are generated. In this paper, we use authorship verification and AI detection as two applications to show how \gtv can be used. For authorship verification, we use the features from \gtv to explain why a pair of documents is by the same or by different authors. We also demonstrate how \gtv features can be used to train a classifier for AI detection, outperforming machine learning models trained on a comparable set of Biber features.
 \\ \newline  }

\begin{document}

\maketitleabstract

\section{Introduction}

Much progress in natural language processing (NLP) has come about from representing linguistic input as vectors of numbers.  These vectors can represent morphological, syntactic, and semantic information.  However, these vectors are not immediately interpretable; for many applications, interpretability is not an issue, but for some applications, it is important to be able to explain to the user why the NLP system has come to a certain conclusion. This is true for high-stakes applications for which system performance is not close to perfect, and for which humans, therefore, remain in the loop.  Examples of such applications include tasks in law enforcement, intelligence, and the judiciary, where wrong decisions may cause harm.
If a human remains in the loop and uses the AI system as a helper, the AI system needs to be able to explain its reasoning.  Explainable AI is a large area of research.  We propose that for some applications, a characterization of linguistic properties, but not of the content, may be helpful.
In this paper, we present our system \gtv, which maps a sentence or document into a sequence of numbers, each of which is interpretable as a specific lexical, morphological, or syntactic feature.  When this text representation is used to make a prediction in a larger system, the system's output can be explained to a user. We use authorship verification and AI detection as two example applications. We plan on releasing our code.

\subsection{Goals of \gtv}
\label{sec:goals}

\gtv is motivated by the goal of creating an interpretable linguistic representation of an author's style. This system aims to provide a more transparent 
approach to text analysis, for example in the context of authorship attribution and stylistic analysis. Specifically, it can provide insights into an author's writing style, such as their use of specific part-of-speech  or sentence patterns, so that a system that uses \gtv can explain its conclusions.

\gtv is designed in a modular manner, so that new groups of features can be added (for example, in certain applications certain choices in content words could be added as a stylistic rather than content indicator).  
Furthermore, existing groups of features can be expanded. For example, new constructions can be added to the set of regular expression testing for specific syntactic constructions. 

The paper is organized as follows. A survey of previous relevant work is provided in Section~\ref{sec:related}. We describe in detail the various feature components of \gtv in Section~\ref{sec:components} and explain the implementation of the syntactic constructions in Section~\ref{sec:srm}. The adaptation of \gtv to Russian is found in Section~\ref{sec:russian}. This is followed by a discussion of our two applications of \gtv. The first is authorship verification, which can be found in Section~\ref{sec:av}, and a case study of \gtv in usage on this task is in Section~\ref{sec:case}. The second application is AI detection, illustrated in Section~\ref{sec:ai}. Finally, we conclude and provide a discussion and implications of our system in Section~\ref{sec:conclusion}.


\section{Related Work}

\label{sec:related}

One system that provides a direct comparison to \gtv is Efficient Linguistic Feature Extraction for Natural Language Datasets, or ELFEN \citep{maurer-2025-elfen}. ELFEN is a Python package for extracting linguistic features from text datasets at scale. While there are a few overlapping feature areas between \gtv and ELFEN, ELFEN contains several unique feature categories including emotion, entities, lexical richness, readability, semantic, and surface feature areas. The full feature list can be found at the GitHub\footnote{\url{https://github.com/mmmaurer/elfen/blob/main/features.md}}. These extracted features can be used for a deeper analysis of text data and NLP model outputs. In this work, we use it as an alternative to \gtv in the authorship verification to see how a comparable interpretable system performs in this task.

Authorship analysis and authorship verification are a set of tasks relevant to many parties. Forensic linguists use authorship verification techniques to narrow down suspects, identify anonymous social media accounts, and provide evidence in court \citep{10.1093/llc/fqad061, Weerasinghe_Singh_Greenstadt_2022, shuy1996language,introtoforensic2016}. Various interpretable methods of authorship verification make use of stylometric features \citep{Stamatatos2016AuthorshipVA} to train classifiers. Some examples include lexical features such as vocabulary, lexical patterns \citep{mendenhall-1887, van-halteren-2004-linguistic}, and syntactic rules \citep{varela2016computational}. These approaches may be somewhat limited because they only capture one, or a few, dimensions of authorship.

For AI detection, the majority of approaches are neural, with a number of commercial detectors as options such as GPTZero \citep{tian2025gptzero} and Pangram \citep{emi2024technical}. A number of studies have shown that Pangram reigns as the most robust of these existing commercial detectors \citep{jabarian2025artificial, russell2025people, chakrabarty2025aiOverWriters}, achieving the lowest false positive and false negative rates.  As for non-neural approaches, our methodology and results for AI detection are directly comparable to those published in \citep{reinhart2025llms} as we use the same corpus with the same goal. The task outlined by \citet{reinhart2025llms} relies on Douglas Biber's tagset of 66 linguistic features \citep{biber1995dimensions, biber1988variation, biber2019register}. While there is some overlap between the Biber features and \gtv features, such as average length metrics and certain phrase structures, there are many differences as well. Most notably, the Biber feature set is significantly smaller and is more oriented towards morpho-syntactic and semantic features.


\section{Components of \gtv}
\label{sec:components}

\subsection{Features}

\gtv consists of nine groups of features.
They are shown in Table \ref{tab:feats} and are described in more detail in the following paragraphs.  Each group is implemented by its own ``featurizer'' which is code that returns a vector of the length shown in Table~\ref{tab:feats}.  All entries are based on counts normalized by the count of its respective feature type.
The final vector returned by \gtv{} is a concatenation of the vectors returned by each featurizer. 

In \gtv, all morphological, POS, and syntactic tags and labels are obtained from the spaCy parser \citep{honnibal2020spacy}.  \gtv runs it once for the document and then obtains the relevant information for each featurizer from the saved output. In actual usage, one can z-normalize a set of \gtv vectors against a background corpus to determine how any one specific author's vector stands out compared to the average author in a dataset.

\begin{table}[h]
\centering
\begin{tabular}{|l|l|}
\hline
Featurizer     & Length \\
\hline
Punctuation marks   & 20     \\
Emojis         & 120 \\
POS Unigrams   & 18     \\
POS Bigrams    & 324     \\
Morphology tags & 46 \\
Dependency labels & 45 \\
Constructions  & 10 \\
Function words & 239    \\
Transition words & 114 \\
\hline
\end{tabular}
\caption{Counts of Implemented Features per Category}
\label{tab:feats}
\end{table}

\paragraph{Punctuation marks}  This featurizer counts the occurrences of various punctuation marks.

\paragraph{Emojis} This featurizer counts over 100 commonly occurring emojis. We also include an out-of-vocabulary emoji feature if we encounter one not in the list.  We chose not to include all emojis, as there are currently 3,953 emojis, most of which are not commonly used.  However, it is easy to add more emojis to this featurizer (and also to remove emojis).


\paragraph{POS tag unigram} This featurizer returns a vector in which each element corresponds to a different tag in the Universal Dependencies tagset \cite{UD}. 

\paragraph{POS tag bigram} This featurizer counts all POS bigrams in a document.  

\paragraph{Morphology tags} These are tags from the UD morphology tagset \cite{tsarfaty-2013-unified} relevant to English, which of course has limited morphology.  For nouns, the features include Case, Definite, and Number; for verbs, Aspect, Number, Person, Tense, and Verbform; and for adjectives, Degree.  

\paragraph{Dependency labels} This featurizer counts all dependency labels in the spaCy dependency parse.

\paragraph{Constructions} This featurizer analyzes the dependency parse tree and detects specific syntactic constructions.  It uses the Syntax Regex Matcher (SRM), which we describe in Section~\ref{sec:srm}. Currently, the analyzed constructions include:
{\em if-because}-clefts, {\em it}-clefts, pseudo-clefts, {\em there}-clefts; subject relative clauses, object relative clauses; yes/no questions, {\em wh}-questions, tag questions; passive voice clauses; and parentheticals.  We have been extending this set for English, and we will include the new constructions in the final version of this paper.

\paragraph{Function words}  This featurizer counts the occurrences of specific function words according to a modified version of NLTK's stopwords list \cite{bird-loper-2004-nltk}. 

\paragraph{Transition words} This featurizer counts the occurrences of sentence-initial transition words in the text. The set of transition words (created by us) includes common conjunctive adverbs, prepositional phrases that act as discourse markers, and other adverbial idioms. 

\section{Syntax Regex Matching}
\label{sec:srm}
The goal of the syntax regex matcher is to detect syntactic constructions, which are defined as patterns against a syntactic parse tree.  It uses the spaCy parse created by \gtv.  SRM transforms the parse tree into an equivalent parenthesized string representation.  For each node of the dependency tree, this representation includes the word, its lemma, its Penn Treebank POS tag (in order to access morphological features), and the arc label connecting the node to its parent in the tree.  We then define regular expressions (in Python) which match against the linearized parenthesized tree representation.  For example, for {\em It was a moral debt that I had inherited from my mother.} we get the following string, which we indent for readability, but which internally has no white spaces.

\begin{verbatim}
(was-be-VBD-ROOT
  (It-it-PRP-nsubj)
  (debt-debt-NN-attr
    (a-a-DT-det)
    (moral-moral-JJ-amod)
    (inherited-inherit-VBN-relcl
      (that-that-WDT-dobj)
      (I-I-PRP-nsubj)
      (had-have-VBD-aux)
      (from-from-IN-prep
        (mother-mother-NN-pobj
          (my-my-PRP$-poss)
          (.-.-.-punct))))))
\end{verbatim}

The regular expressions then match against these strings.  For example, the regex for the {\it}-cleft looks like this:
-
\begin{verbatim}
\([^-]*-be-[^-]*-[^-]*
  .*\([iI]t-it-PRP-nsubj\)
  .*\([^-]*-[^-]*-NNP?S?[^-]*-attr
    .*\([^-]*-[^-]*-VB[^-]*-
    (relcl|advcl)
\end{verbatim}

It searches for a form of {\em to be} which has, among other dependents, a subject whose lemma is {\em it}, and a nominal dependent with the {\tt attr} relation, and with a relative clause.  We note that the {\tt attr} is not part of the UD set of relations, and syntacticians may dispute the analysis spaCy produces for {\em it}-clefts, but we of course base our regular expression on what spaCy actually produces for standard {\em it}-cleft sentences, rather than on what we believe it should produce.

We note that there may be constructions such that it is impossible to find all and only instances of the constructions in a set of arbitrary trees using our approach, but that in practice this is not a problem because the parse trees have, by and large, predictable shapes.

\section{Adapting \gtv to Russian}
\label{sec:russian}
We apply the same approach to Russian, incorporating both universal and language-specific morphosyntactic features. While universal features (e.g., POS unigrams and bigrams, punctuation, emojis) were retained, language-specific ones (functional words and morphosyntactic configurations) were adjusted to capture structural differences that are unique to Russian.

Even though some syntactic patterns are shared with English, their dependency configurations vary due to different surface structures, dependency labels, and morphological tags. Thus, these similarities do not have exact one-to-one correspondences. For example, some constructions such as the passive voice exist in both languages, but are realized differently in terms of syntactic dependencies: in Russian, it can be formed not only with an auxilary but also with a reflexive verb. On the other hand, some constructions are language-specific. For instance, English \textit{it-clefts} have no equivalent in Russian, and Russian also has unique constructions absent in English. Therefore, we adjusted the set of regular expressions for Russian by adapting shared constructions and introducing new, language-specific patterns.

One example of a Russian-specific feature we incorporated is participles. In Russian, participles (similar to gerunds) often serve as stylistic markers of certain authors and genres, as they are relatively rare in everyday speech. Russian has four types of participles, distinguished by tense (past vs. present) and voice (active vs. passive). Although SpaCy’s morphological tag \texttt{VerbForm=Part} does exist for Russian (unlike English), it alone does not reliably detect all participle forms. Moreover, the same tag is assigned to gerunds, but we aimed to differentiate participles from gerunds to increase precision. Therefore, we incorporated specific suffixes used in participle formation into the regular expression. Below is an example of a regular expression designed to capture constructions containing participles of all four types (present/past and active/passive):


\newcommand{\rus}[1]{\foreignlanguage{russian}{#1}}

\begin{alltt}
\begin{small}
\textbackslash{}(\textbackslash{}([\textasciicircum-]*-[\textasciicircum-]*-(?:VERB|ADJ)-ROOT.
 *?(?:ADJ|VERB)-(?:amod|acl)(?!:relcl)
 .*?(?:VerbForm=Part|[\rus{а-я}]+  
 (?:\rus{ющ}|\rus{ащ}|\rus{ящ}|\rus{вш}|\rus{ем}|\rus{им}|\rus{нн}|\rus{т})[\rus{а-я}]+).*?\textbackslash{})+\$)
\end{small}
\end{alltt}


In addition to syntactic patterns, we modified morphological features, as prior research has shown that morphological analysis contributes to authorship attribution, in particular in languages with richer morphology than English \cite{Navot2006, 10.1007/978-3-030-37334-4_18}.
Building on this, we expanded the standard set of morphological tags by including those specific to Russian (e.g., the six grammatical cases and animacy). We also added derivational morphological features. For example, we incorporated \textit{diminutives}, which denote small or affectionate objects (e.g., \foreignlanguage{russian}{книжечка} `little book'). We detected these forms by designing a regular expression that reflects morphological rules of the language. Including this feature in our set is consistent with \cite{okulska2023stylometrix}, where diminutives are also used in building StyloMetrix, the open-source multilanguage stylometric tool.

Next, we incorporated a feature for identifying \textit{feminitives}, words derived from masculine forms through special suffixation patterns to denote female professionals (e.g. \foreignlanguage{russian}{авторка} `female author', \foreignlanguage{russian}{\textit{профессорша}} `female professor'). Although prior studies have noted an increasing use of feminitives in contemporary language and examined their word-formation patterns \cite{Nesset2022}, to the best of our knowledge, this feature has not been yet used in rule-based approaches for stylometric or authorship attribution tasks for Russian. However, we believe that the use of feminitives exhibits sociolinguistic variation. While some forms are standard and widely accepted, others represent neologisms that may sound unconventional to native speakers of certain ages or social backgrounds. As an initial step, we used a publicly available list of feminitives from \href{https://ru.wiktionary.org/wiki/%D0%9A%D0%B0%D1%82%D0%B5%D0%B3%D0%BE%D1%80%D0%B8%D1%8F:Nomina_feminina/ru}{Wikipedia}.
In future work, we plan to replace this approach with the rule-based one that allows to distinguish between conventional and neologistic feminitives, enabling more nuanced stylistic analysis. 

\section{Application: Authorship Verification}
\label{sec:av}
Our first example application for using \gtv is authorship verification (AV). Authorship verification is a task with the goal of determining how likely a pair of documents share the same author. 

\subsection{Datasets}
For the task of authorship verification, we select three datasets used by \citet{rivera-soto-etal-2021-learning} in their work creating LUAR, which is a neural model that has state-of-the-art performance in authorship tasks. (We do not compare to LUAR in this paper because we focus on interpretable features, and LUAR's features are not interpretable.) These three datasets, Reddit comments, Amazon reviews, and Fanfiction stories, cover a diverse set of genres, and offer insight on how systems perform in a variety of domains.

\subsection{Method and Metrics}

The authorship verification task is fundamentally a similarity task: we use \gtv to turn pairs of documents into high dimensional embeddings, and then we use the cosine similarity to measure the distance between these pairs of embeddings. To obtain a specific answer (same author/different author), the cosine similarity is compared to a tuned threshold which thus determines the prediction. 

\subsection{A Comparison of Interpretable Document Vectorizers}

We investigate Efficient Linguistic Feature Extraction for Natural Language Datasets, or ELFEN \cite{maurer-2025-elfen}, another interpretable document vectorizer. While there are some overlapping features between \gtv and ELFEN, ELFEN contains several unique categories of features distinct from those in \gtv; see Section~\ref{sec:related} for details. 

\input{tables/elfen}
We show the results of using both \gtv and ELFEN on our datasets in Table~\ref{tab:elfen}. While both systems perform comparably on the three datasets (with \gtv performing better on two datasets, though we do not do significance testing and do not claim this as a result), what's interesting is that concatenating the two feature sets consistently does as well as or somewhat better than either system alone. This indicates that these two featurizers may capture some orthogonal information.

\input{figures/case-study}

\subsection{Russian \gtv for Authorship Verification}

We also perform a preliminary experiment using our adapted Russian version of \gtv on a dataset consisting of Pikabu comments from \citet{ilya_gusev_2024} available on HuggingFace. By evaluating \gtv on the Russian Pikabu dataset, we get an AUC of 0.65, which is comparable to our results using \gtv on the various English datasets. 

\section{\gtv for Authorship Verification Case Study}
\label{sec:case}

We present two examples of authorship verification to show how \gtv can be used to predict authorship, and how its features can be used to explain the prediction. We can show which features were most impactful in the prediction, and to what extent they determined the prediction. 

We define a criterion to select these most impactful features, depending on whether a pair of documents are predicted to be by the same or different authors. When a pair of documents is predicted to be written by the same author, we want to maximize the absolute values of the feature values (features that distinguish these documents from the large set of background documents) while making sure the values are similar for both documents. Thus, for identifying features for same-author pairs, we use $|val\_1| + |val\_2| - |val\_1 - val\_2|$, where $val\_1$ represents the feature's score for document 1, and $val\_2$ represents the feature's score for document 2. When a pair of documents is predicted to be by different authors, we simply find the largest magnitudes of differences in the feature values, i.e., we use $|val\_1 - val\_2|$.  We then choose the top $n$ features; in the examples below, we use $n=10$.

\subsection{Example 1: Different Author Pair}
\label{sec:example_1}

Looking at the first example in Figure \ref{fig:example_pairs}, 
and sort in descending order the top 10 features. These represent the 10 most differing features in the pair of documents.  Looking at the features, we first note several function words which can be found in document 2 but not in document 1; for example, document 2 uses {\em when} twice in a fairly short text, while  document 1 does not use it at all.  In contrast, document 1 uses several part-of-speech (POS) bigrams far more frequently than the background corpus, while document 2's distribution of POS bigrams is more standard.  A striking example is the bigram adjective-proper noun, which is unusual in general but very frequent in document 1 ({\em vengeful Sith}, {\em fallen Master}, {\em former Senator}, {\em wounded Padawan}).  Finally, we note the high frequency of the indefinite article in document 1: {\em a scythe}, {\em a new darkness}, {\em a cold, lonely mist}, {\em a dismembered former senator}, {\em a shorn and wounded Padawan}, {\em a Jedi Master}.  These indefinite noun phrases provide a sense of change (indefinites introduce new discourse objects); in the case of the last three, the author takes on the perspective of three characters. Document 2, in contrast, has few indefinites and the narration centers on entities known to the readers and the characters in the story.

\input{tables/different-author-scores}
\subsection{Example 2: Same Author Pair}

In this case, the cosine similarity between the two \gtv embeddings is 0.20, so based on our threshold of 0.15, we determine that \gtv predicts that the two documents are written by the same author, which is in fact the case.  When identifying similar features in two documents, we use the metric $|val\_1| + |val\_2| - |val\_1 - val\_2|$ and take the top 10 features in descending order, shown in Table~\ref{tab:same_author_scores}.
Thus, these are features which occur in both documents either much more or much less frequently than on average across a background corpus.  One example is the bigram preposition-punctuation.  In both texts, we find examples: {\em UP!}, {\em up!} (in document 1), {\em out.}, {\em of.} (in document 2).  A preposition at the end of a clause is often discouraged in formal written English.   
The two documents also use passive voice clauses more frequently than on average (passive voice is generally rare in written English): {\em it was connected} (document 1), {\em it was passed down}, {\em it was created} (document 2).  
The two documents share a negative value for the punctuation mark comma.  Indeed, neither text contains a comma, which in general is a very common punctuation mark.

\section{Application: AI Detection}
\label{sec:ai}
Our second example application for \gtv is AI detection. The goal is to determine whether Large Language Models (LLMs) replicate human writing styles sufficiently well to avoid detection, or whether the distribution of features between text written by a given LLM and a human is disparate enough to allow correct classification. Logistic regression is used to measure prediction accuracy, and the performance of \gtv features for this task is compared to previous work.

\input{tables/same-author-scores}

\subsection{Datasets}
The motivation behind this task is the intuition that it is sometimes possible to determine that a text was written by an LLM due to over-use of certain features such as punctuation types, or an overall ``unnatural'' writing structure. We use \gtv to more formally encode these features in an attempt to determine the legitimacy and applicability of these otherwise anecdotal claims. 

\input{tables/hape_to_cap}

To test the applicability of \gtv features for AI detection, we use the Human-AI Parallel English Corpus (HAP-E) and the COCA AI Parallel Corpus (CAP), created by \citet{reinhart2025llms}. The HAP-E corpus was constructed by splitting 1,000-word samples of human-written texts into two chunks. The first chunk of each sample was then fed to various LLMs, which were prompted to produce another 500 words approximately on the same topic in the same style and tone as the original. The corpus includes these LLM-generated texts and the human-written counterparts, and it spans a wide range of genres such as academic articles, news articles, fictional stories, podcast transcriptions, blogs, and, movie scripts. Distributions of genres were selected to approximate the composition of CAP. The LLMs used in the creation of this corpus are GPT-4o, GPT-4o-mini, base and Instruct versions of Llama-3-8B, and base and Instruct versions of Llama-3-70B. For each author type, there is a total of 8,290 entries in HAP-E and 9,615 entries in CAP, and a given index refers to a parallel writing sample across author files. For the purpose of this task, the second half of the human-written texts (chunk\_2) were compared to the texts generated by each LLM individually. 

The goal is to determine whether LLMs are able to effectively replicate human writing styles in terms of feature frequencies. The use of an interpretable model also reveals which features are most predictive, which can be used to fine-tune which \gtv features are used as a way to reduce run-time costs or to tailor \gtv models for specific tasks.\footnote{In principle, the ability to see which features are most predictive could also be exploited for tasks such as improving AI performance in style matching.}

\subsection{Method and Metric}

We use the logistic regression model from scikit-learn \citep{scikit-learn} on the data and train to weight the \gtv features. We use logistic regression because the weights provide highly interpretable results which can be used to explain the reasoning of the system. Weights were determined by training the model on the HAP-E corpus, and these were tested on the non-overlapping CAP corpus to see how generalizable the results are.


To evaluate the effectiveness of our method, we simply use accuracy: how often do we determine the correct author.  Note that the random baseline is 50\%.

\subsection{Results}
Our results reveal that \gtv is highly effective for identifying stylistic differences between human writing samples and LLM text generations.

An initial experiment used a set of hand-picked features that seemed relevant to the task of AI detection. These features include word count, unique words (type count), average sentence length by token, transition word count, and unique transitions count. With just these features, the model yielded 93\% accuracy on the GPT-4o test set.

Using the full set of \gtv features, we achieve over 98\% accuracy on two thirds of the data, which is the highest overall performance compared to the results reported by \citet{reinhart2025llms}. These comparisons and results are summarized in Table~\ref{tab:hape_to_cap}.

However, because \gtv includes hundreds of features, it can be difficult to make strong claims about how to interpret the role of individual features in the AI detection task. Since the primary goal is interpretability, we repeated the experiment using only the top 10 most highly weighted features, which are listed along with their weights in Table~\ref{tab:top10}. On this smaller subset of features, the model still achieved 95\% accuracy on human versus GPT-4o data. Reducing the set of features allows for a more straightforward analysis of the defining differences between the writing styles of two author types, which may be useful for tasks such as plagiarism detection. For example, the highest weighted feature is type count, and the negative value indicates that GPT-4o uses a more varied vocabulary than human authors. Punctuation such as periods and semicolons are used more often by humans, which may suggest that humans segment their writing into sentence or independent sentence-like clauses more often than GPT-4o. Additionally, the weights associated with the (ADJ, ADJ) POS bigram and the exclamation point suggest that humans samples are more descriptive or emotive than their GPT-generated counterparts. Given the high performance on these features, we can conclude that these different feature distributions are not statistical anomalies, but are  significant indicators of human versus AI authorship.

\input{tables/top10}

\subsection{A Comparison of Feature-Based AI Detectors}
To understand how competitive \gtv is, we compare its performance to that of the Biber feature-based model \citep{reinhart2025llms}. As previously discussed, the Biber features are not exactly the same as the \gtv features, and \gtv has many hundreds more features. This large quantity of features can be a great advantage for \gtv in achieving high performance. While \citet{reinhart2025llms} do not use the same evaluation metrics, Table~\ref{tab:hape_to_cap} 
shows that even their best results are generally outperformed by those obtained with \gtv. The only exception is the Llama base models classified with a random forest pairwise method, which is primarily due to irregularities in the data caused by differences in LLM architectures. Llama Instruct models and GPT-4 models are similar to each other because they are both instruction-tuned models that are designed to follow user directions, while Llama base models are simpler next-token prediction models. As a result, there is more noise in data from the Llama base models, and the salient features from Llama Instruct and GPT-4 are not as prominent. As a non-linear  model, random forest pairwise classification is better suited to handle the irregularities in Llama base data, despite yielding lower performance overall.

However, recalling the results given in Table~\ref{tab:top10}, even when we limit our trial to the top 10 most prominent \gtv features in the trial, our results are comparable to those achieved by \cite{reinhart2025llms}. The implication is that \gtv features are generally better-suited for the task of AI detection than the Biber features.

Overall, the strong results obtained using \gtv features demonstrate that we have developed a competitive system for AI detection. Even with a significantly reduced subset of features, our system is still comparable to similar approaches. The high performance achieved on the full feature set, coupled with the fact that the results are transparent and directly interpretable, set \gtv comfortably in the ranks of effective tools for feature-based text analysis.

\section{Conclusion}
\label{sec:conclusion}

In this paper, we present \gtv, a grammatical style embedding algorithm that embeds an author's linguistic choices into higher dimensional space, at the same time remaining inherently and immediately interpretable. The \gtv system is modular, allowing for the addition of new groups of features and the expansion of existing ones. It can also be adapted to other languages besides English. We adapt \gtv to Russian, adding additional features and Russian-specific constructions. For authorship verification, we achieve comparable results on Russian authorship verification to English authorship verification. We show how \gtv can be used to explain an authorship prediction using its interpretable features.

We also illustrate how to use \gtv for AI detection. For AI detection, \gtv yields SOTA results while being fully explainable. Even on a small subset of features, \gtv achieves performance that is comparable to other feature-based AI detectors. We believe this approach to be a promising direction towards developing more interpretable NLP systems. Our \gtv system will be freely available (URL in final version of paper).

\section{Ethical Considerations and Limitations}
We have demonstrated that \gtv can be used to make predictions both for authorship verification and AI detection tasks. The interpretable results mean that the user can point to specific features as evidence for authorship claims. However, we stress that \gtv is a tool that is intended to support human-driven decisions, perhaps in tandem with other data, rather than as a substitution for human discernment. Although \gtv yields high classification accuracy in our experiments, these results are not perfect and should not be cited as exclusive evidence in judicial processes, plagiarism accusations, or similar tasks.

While the results of our AI detection experiments are promising, the scope of our analysis is limited due to the small number of datasets used, and the fact that these datasets are parallel as opposed to being naturally-occurring. We plan to run tests on more challenging datasets to see how \gtv compares to commercial AI detectors, bearing in mind the limitations of our neural-free approach.



\nocite{}
\section{Bibliographical References}
\label{sec:reference}

\bibliographystyle{lrec2026-natbib}
\bibliography{lrec2026-example, anthology}

\bibliographystylelanguageresource{lrec2026-natbib}
\bibliographylanguageresource{languageresource}

\end{document}

%% file: tables/elfen.tex
\begin{table}[htb]
\setlength\tabcolsep{4pt}
\centering
\small
\begin{tabular}{lcccc}  
\toprule
\textbf{Data}  & \textbf{\gtv} &  \textbf{ELFEN} & \textbf{Combined}  \\
\midrule
Reddit      & 0.63 & 0.59 & 0.63 \\
Amazon      & 0.71 & 0.71 & 0.74 \\
Fanfiction  & 0.69 & 0.66 & 0.70  \\
\bottomrule
\end{tabular}
\caption{The results evaluating \gtv, ELFEN, and the concatenation of both on the Reddit, Amazon, and Fanfiction datasets. While the performance of both \gtv and ELFEN are comparable, the concatenation of the two tends to perform a bit better than either feature set alone.}
\label{tab:elfen}
\end{table} 

%% file: figures/case-study.tex
\begin{figure*}[ht]
    \begin{tcolorbox}[
        colback=yellow!15!brown!10!white,    
        colframe=brown!20!white!80, 
        arc=5mm,
        width=\textwidth,
        before upper={\parindent15pt},
    ]
    \small
    \noindent\textbf{Example Pair 1: Different Author}
    \vspace{0.5em}
    
    \noindent\textbf{Document 1:}\\
    Whirling like  \highlight{a} scythe, the saber sliced her upper torso, putting an end to the \highlight{vengeful Sith}. Dropping to her knees again, Jameh crawled to her \highlight{fallen Master}, cradling him in her arms.  \highlight{A} new darkness grew in her heart now, one like a cold, lonely mist. 
    Pilae, Obi-Wan, and Anakin stood nearby, dismayed at the sight that met their eyes:  \highlight{a}  dismembered \highlight{former Senator}, \highlight{a} shorn and \highlight{wounded Padawan}, and  \highlight{a} Jedi Master on the verge of death. " Master, please, you can"t leave me. I need you; I"m not ready!"
    
    \medskip
    \noindent\textbf{Document 2:}\\
    He scanned the field beyond and was dumbfounded \highlight{when} he didn"t see any rats. 
    He pelted back through the entrance and into the clearing. The Clan had been alerted by Redfur"s yowl of surprise, so they had stopped chatting and lowered their bodies into a crouch, getting ready for the rats. But \highlight{when} they saw the four rats clinging to Redfur"s fur, they hissed in astonishment at the size of them.
    \vspace{0.5em}
    
    \noindent\textbf{\gtv Cosine Similarity}: 0.09,
    \vspace{0.5em}
    \hrule
    \vspace{0.5em}
    
    \noindent\textbf{ Example Pair 2: Same Author}
    \vspace{0.5em}
    
    \noindent\textbf{Document 1:}\\
    GET \highlight{UP!} School time!" Sora called from the door. " I"m \highlight{up!}" he hollered back before throwing the cover"s off him. It"s been a week. A week since Roxas started hearing that voice. Throughout that time he had figured out that \highlight{it was connected} to the mirror he had gotten at the same time. "
    
    \medskip
    \noindent\textbf{Document 2:}\\
    \highlight{It was passed down} through generations to keep him in the glass." At this he closed the book and plopped on the bed. " What about the rhyme?" Demyx stroked his chin in a pondering position. " \highlight{It was created} to scare children from letting him out. Though the ending part. "" A curse to never be free \highlight{of.} Until this demon admits love" Is exactly what it says.

    \vspace{0.5em}
    \noindent\textbf{\gtv Cosine Similarity}: 0.20, 
    \end{tcolorbox}
    \caption{Example Pairs for Case Study. Pair 1 is by two different authors, and Pair 2 is by the same author.}
    \label{fig:example_pairs}
\end{figure*}

%% file: tables/different-author-scores.tex
\begin{table}[htbp]
    \centering
    \small
    \begin{tabular}{@{}l@{\hspace{3pt}}r@{\hspace{3pt}}r@{\hspace{3pt}}r@{}}
    \toprule
    \textbf{Feature} & \textbf{Score} & \textbf{Doc 1} & \textbf{Doc 2} \\
    \midrule
    func\_words:further & 5.4 & -0.1 & 5.3 \\
    pos\_bigrams:ADJ PROPN & 4.1 & 3.8 & -0.3 \\
    pos\_bigrams:PUNCT DET & 3.8 & 3.4 & -0.4 \\
    func\_words:through & 3.6 & -0.3 & 3.3 \\
    pos\_bigrams:PART ADJ & 3.3 & 3.1 & -0.2 \\
    func\_words:they & 2.9 & -0.4 & 2.5 \\
    pos\_bigrams:PROPN PUNCT & 2.9 & 2.1 & -0.8 \\
    morph\_tags:Definite=Ind & 2.8 & 2.4 & -0.4 \\
    pos\_bigrams:PUNCT NUM & 2.6 & 2.5 & -0.1 \\
    func\_words:when & 2.6 & -0.4 & 2.2 \\
    \bottomrule
    \end{tabular}
    \caption{Feature scores comparison between Example 1 document pair by different authors.}
    \label{tab:different_author_scores}
\end{table}

%% file: tables/same-author-scores.tex
\begin{table}[htbp]
    \centering
    \small
    \begin{tabular}{@{}l@{\hspace{3pt}}r@{\hspace{3pt}}r@{\hspace{3pt}}r@{}}
    \toprule
    \textbf{Feature} & \textbf{Score} & \textbf{Doc 1} & \textbf{Doc 2} \\
    \midrule
    pos\_bigrams:PREP PUNCT & 6.1 & 4.1 & 3.0 \\
    passive sentence & 5.4 & 2.7 & 4.6 \\
    dep\_labels:nsubjpass & 4.4 & 2.2 & 3.8 \\
    pos\_bigrams:PREP VERB & 4.3 & 2.9 & 2.1 \\
    dep\_labels:auxpass & 3.7 & 1.8 & 3.3 \\
    func\_words:from & 3.5 & 2.3 & 1.8 \\
    punctuation:, & 3.4 & -1.7 & -1.7 \\
    morph\_tags:PunctType=Comm & 3.3 & -1.6 & -1.6 \\
    pos\_bigrams:DET NOUN & 3.2 & 2.5 & 1.6 \\
    func\_words:the & 2.9 & 1.5 & 1.5 \\
    \bottomrule
    \end{tabular}
    \caption{Feature scores comparison between Example 2 document pair by the same author.}
    \label{tab:same_author_scores}
\end{table}

%% file: tables/hape_to_cap.tex
\begin{table*}[t]
\centering
\begin{tabular}{@{}lccc@{}}
\toprule
& \textbf{\gtv} & \multicolumn{2}{c}{\textbf{Biber Features}} \\
LLM & Logistic Regression & Random Forest Pairwise & Lasso Pairwise \\
\midrule
\multicolumn{3}{@{}l}{\textit{GPT-4o}} \\
\quad GPT-4o & \textbf{99.3\%} & 98.4\% & 95.3\% \\
\quad GPT-4o Mini & \textbf{99.6\%} & 98.2\% & 96.2\% \\
\addlinespace
\multicolumn{3}{@{}l}{\textit{Llama Instruct}} \\
\quad Llama 3 70B Instruct & \textbf{98.1\%} & 93.5\% & 90.2\% \\
\quad Llama 3 8B Instruct & \textbf{98.1\%} & 95.3\% & 91.9\% \\
\addlinespace
\multicolumn{3}{@{}l}{\textit{Llama Base}} \\
\quad Llama 3 70B & 89.1\% & \textbf{94.6\%} & 75.3\% \\
\quad Llama 3 8B & 91.2\% & \textbf{93.6\%} & 75.3\% \\
\bottomrule
\end{tabular}
\caption{Comparison of accuracy from \gtv and \citet{reinhart2025llms} trials when trained on HAP-E and tested on CAP.}
\label{tab:hape_to_cap}
\end{table*}

%% file: tables/top10.tex
\begin{table}[htb]
\setlength\tabcolsep{4pt}
\centering
\small
\begin{tabular}{lccc}  
\toprule
\textbf{Feature} & \textbf{Type} & \textbf{Weight}\\
\midrule
types           & count & -1.051 \\
punctuation     & . & 0.532 \\
dep\_labels      & advcl & -0.475 \\
punctuation     & : & -0.474 \\
pos\_bigrams     & ADJ ADJ & 0.463 \\
punctuation     & - & 0.458 \\
punctuation     & ; & 0.449 \\
func\_words      & which & 0.435 \\
punctuation     & ! & 0.429 \\
morph\_tags      & Aspect=Prog & -0.425 \\
\bottomrule
\end{tabular}
\caption{Top 10 features used in human vs. GPT-4o detection on HAP-E, in decreasing order of predictive power. Positive weights indicate correlation with human writing style, while negative weights indicate correlation with GPT-4o writing.}
\label{tab:top10}
\end{table}